\documentclass[letterpaper, 10 pt, conference]{ieeeconf}  

\IEEEoverridecommandlockouts                              

\usepackage[shortcuts,acronym]{glossaries}

\newacronym{ah}{AH}{Adaptive Height}
\newacronym{fh}{FH}{Fixed Height}
\newacronym{rcs}{RCS}{Radar Cross Section}
\newacronym{map}{mAP}{mean Average Precision}
\newacronym{fpn}{FPN}{Feature Pyramid Network}
\newacronym{ehl}{EHL}{Enhanced Huber Loss}
\newacronym{fmcw}{FMCW}{Frequency-Modulated Continuous-Wave}
\usepackage{times}
\usepackage{epsfig}
\usepackage{graphicx}
\newsavebox\mybox
\usepackage{amsmath}
\usepackage{amssymb}
\usepackage{subcaption}
\usepackage{amsmath}
\usepackage{multirow}
\usepackage{float}
\usepackage{xspace}
\usepackage[flushleft]{threeparttable}
\usepackage[compress]{cite}
\usepackage[font=small,labelfont=bf]{caption} 

\author{Huawei Sun$^{1,2}$, Hao Feng$^{2}$, Gianfranco Mauro$^{1}$, Julius Ott$^{1,2}$, Georg Stettinger$^{1}$, Lorenzo Servadei$^{2}$, Robert Wille$^{2}$
\thanks{$^{1}$Infineon Technologies AG, Neubiberg, Germany
        {\tt\small \{huawei.sun, gianfranco.mauro, julius.ott, georg.stettinger\}@infineon.com}}%
\thanks{$^{2}$Technical University of Munich, Munich, Germany
        {\tt\small \{hao.feng, lorenzo.servadei, robert.wille\}@tum.de}}%
}

\begin{document}
\title{\LARGE \bf
Enhanced Radar Perception via Multi-Task Learning: Towards Refined Data for Sensor Fusion Applications
}



\maketitle
\thispagestyle{empty}
\pagestyle{empty}

\begin{abstract}
Radar and camera fusion yields robustness in perception tasks by leveraging the strength of both sensors. The typical extracted radar point cloud is 2D without height information due to insufficient antennas along the elevation axis, which challenges the network performance. This work introduces a learning-based approach to infer the height of radar points associated with 3D objects. A novel robust regression loss is introduced to address the sparse target challenge. In addition, a multi-task training strategy is employed, emphasizing important features. The average radar absolute height error decreases from 1.69 to 0.25 meters compared to the state-of-the-art height extension method. The estimated target height values are used to preprocess and enrich radar data for downstream perception tasks. Integrating this refined radar information further enhances the performance of existing radar camera fusion models for object detection and depth estimation tasks. 

\end{abstract}

\vspace{-1.2mm}

\section{Introduction}
In autonomous driving, information from multiple sensors such as camera, lidar, and radar is usually fused to compensate for the shortcomings of each single sensor \cite{survey}. Thus, fusion-based approaches usually perform better in accomplishing perception tasks \cite{multi1,multi2,multi3,qi2018frustum}.
Radar sensors alone present several limitations. The generated radar point clouds tend to be considerably sparser than those generated by lidar. Further, many of the radar sensors employed in public datasets, such as the nuScenes \cite{nuscenes}, PixSet \cite{deziel2021pixset},
suffer from poor spatial elevation resolution. This aspect leads to absent height information for the radar points. 

To effectively address perception tasks such as object detection, it is essential to preprocess the radar data before integrating it with visual images. Several contributions, such as  \cite{crf,MCAF} extend the height value of each radar point before projecting it onto the image plane. This method visually represents each radar point as a vertical line on the image plane, substantially increasing the density of radar projection maps. However, these methods have not considered the association between the radar points and the objects.
It is necessary to consider the ground truth vertical dimension of the corresponding object size to precisely determine the extended height values of individual radar points. This necessitates undertaking a task known as \emph{target height estimation}. Numerous studies have endeavored to estimate object height utilizing \acrfull{fmcw} radar signals. However, these studies either utilize traditional signal processing methods \cite{height_curb,height_entrance} or employ random forest regressors \cite{height_obj_randomforest,height_humanbody} to estimate the height of a single object over a series of consecutive frames.

In contrast, our experimental setup introduces greater complexity by aiming to ascertain the height of every radar point associated with various objects within each frame. This challenge is compounded by the variability in the number of objects and radar points from frame to frame. Consequently, neither conventional signal processing methodologies nor random forest algorithms are equipped to address our problem. Moreover, it is essential to highlight a fundamental distinction in our objectives compared to the contributions cited above. While the listed papers are primarily concerned with the singular task of height estimation for individual targets, we aim to enhance radar data quality, ultimately serving as a valuable input for downstream perception tasks.

Thus, this study presents a learning-based approach to ascertain the height of radar points via predicting a height map within the image plane. For the pixel coordinates where radar points are located, the predicted height values serve as a reliable measure of the vertical extension values that correlate with the radar points. A robust regression loss is proposed to address the challenges of sparse target regression. To mitigate the issue of the predicted height map reverting to all-zero values, we employ a multi-task training strategy, estimating height and segmenting free space simultaneously. This method substantially decreases the average radar absolute height error from 1.69 to 0.25 meters, offering a notable improvement over the \acrfull{ah} approach \cite{MCAF}. The estimated height values can serve as definitive extensions to refine the radar data and accomplish subsequent perception tasks. Incorporating the deduced height values for preprocessing, the \acrfull{map} of the MCAF-Net \cite{MCAF} and CRF-Net \cite{crf} increases. Meanwhile, the DORN \cite{Dorn_radar} algorithm also performs better when applying our learning-based height extension. This highlights the crucial role of precise radar data in enhancing the performance of perception tasks. To our knowledge, this is the first method employing deep learning for radar point height derivation, establishing a new benchmark in radar data quality.
\label{sec:intro}
\vspace{-1.2mm}

\section{Related Work}
\subsection{Radar Height Estimation}
In the work of \cite{height_curb} and \cite{height_entrance}, the Doppler effect-induced frequency shift is exploited to estimate the height of a curb and an entrance gate, respectively. Meanwhile, \cite{height_curb_brid} derives a target height formula founded upon the geometric relationship between the sensor and target, facilitating the estimation of curb and brick heights in real-world scenarios. The studies \cite{height_corner} and \cite{height_vehicle} put forward a novel approach involving the detection of corner reflectors positioned at varying heights by analyzing radar wave multipath propagation. \cite{height_classification} make use of properties of direct and indirect reflections between a radar sensor and an object to infer the height. This height information, along with parameters like vehicle velocity and radar sensor mounting height, is integrated into a classification algorithm to categorize the target as either traversable or non-traversable. Departing from traditional signal processing techniques, \cite{height_obj_randomforest} and \cite{height_humanbody} employ random forest regressors to estimate object and human body heights, respectively. 
However, these approaches are tailored for single-target scenarios and not appropriate for the complex, multi-target dynamic settings encountered in autonomous driving.

\subsection{Radar Data Refinement in Autonomous Driving Applications}
Processed radar point clouds retrieved from public datasets like nuScenes often lack height information. 
In \cite{centerfusion}, the authors expand each radar point into a fixed-size pillar before associating the radar detections with the image features of their corresponding objects.
This technique enhances the initial object detections within the 3D space.
Similarly, in \cite{crf, fpp}, the authors employ the \acrfull{fh} method to extend the height value of each radar point to a predetermined fixed value before projecting it onto the image plane. A more sophisticated approach known as \acrfull{ah} extension is introduced in \cite{MCAF}. This method accounts for factors such as \acrfull{rcs} and the distance characteristics of individual radar points to determine the appropriate extension value precisely. Consequently, this approach ensures a highly precise alignment of radar data with the corresponding objects in the scene.

Regarding radar-camera fusion for depth completion tasks, 
the DORN algorithm presented in \cite{Dorn_radar} follows the extension approach in \cite{crf} by expanding the height of radar points within a range of 0.25 meters to 2 meters. Each point is projected onto the image plane as a vertical line, carrying valuable depth information.

However, these previous studies overlooked the crucial aspect of the relationship between radar points and the objects they represent, which resulted in projections marred by noise. In response, our work employs deep learning techniques to deduce the height value of each radar point, considering the connection between the radar points and their corresponding real-world objects.
\label{sec:related_work}
\vspace{-1.2mm}

\section{Problem Formulation}
\begin{figure}[htbp]
\centering
     \includegraphics[width = 0.90\linewidth]{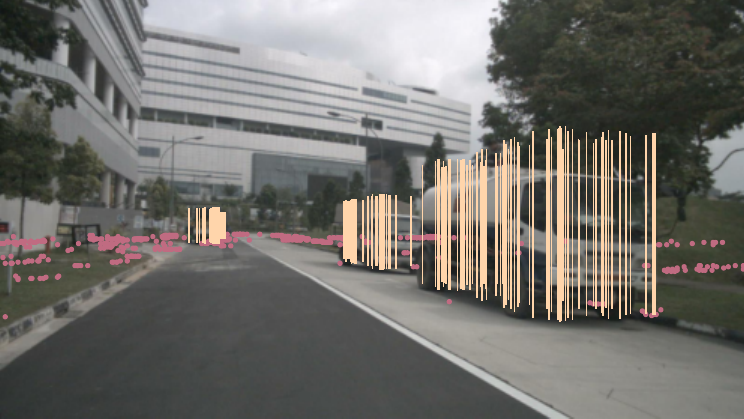}
     \caption{Example of ideal radar projection. Extended vertical lines of radar points correspond to objects are colored in
yellow. Others are marked in red.}
     \label{Fig:ideal_proj}
     \vspace{-4mm}
 \end{figure}
\noindent As demonstrated in \cite{MCAF}, accurate radar information empowers models to distinguish objects better, thereby enhancing the \acrshort{map} in object detection tasks.
This underscores the necessity of attaining precise height estimation for radar points. 
Ideally, for a radar point associated with an object in 3D space, an accurate height prediction ensures that the corresponding vertical line aligns seamlessly with the 2D bounding box in the image. This practically corresponds to the yellow lines of Fig. \ref{Fig:ideal_proj}. Meanwhile, radar points not affiliated with any objects, often viewed as noise, appear as individual red pixel points. 
By setting a specific height threshold, we can effectively discard such outlier radar points, leading to a cleaner and more accurate representation of the radar image. 

In a precise manner, within each frame, we denote by $\mathcal{B}=\{B^{1}, B^{2}, ..., B^{n}\}$ a collection of object bounding boxes, where $n$ objects are present in the current frame. The respective heights of these objects are represented as $\mathcal{H}={H^{1},H^{2}, ..., H^{n}}$. 
For the $k^{th}$ radar projection $R^{k}$ situated at image coordinates $(i,j)$, we establish the ground truth height $H(i,j)$ of this specific radar point through the following formulation:
\vspace{-1mm}

\begin{equation}
    H(i,j) = \left\{ \begin{array}{rcl}
        H^{m} & \mbox{if} & R^{k} \in B^{m} \\
        0 & \mbox{if} & R^{k} \notin \mathcal{B},
    \end{array} \right.
\vspace{-1mm}
\end{equation}
where $m$ denotes the $k^{th}$ radar point is associated with the $m^{th}$ object.

Common regression tasks often employ L1 and L2 loss functions. L2 loss, while advantageous for its rapid convergence, is prone to the influence of outliers \cite{l2}. Conversely, L1 loss assigns linear penalties to errors, which does not disproportionately penalize larger errors as L2 does. Thus, it is less sensitive to outliers but can be less computationally efficient. The Huber loss \cite{huber} bridges the gap between these two, transitioning smoothly between L1 and L2 loss.
Thus, the Huber loss is widely used in regressing the bounding box positions \cite{fast, faster_rcnn}. This provides the benefit of effective network training while reducing sensitivity to outliers. Despite their merits, networks can potentially get trapped in local minima when relying solely on these conventional loss functions in sparse target regression scenarios like ours.

\label{sec:problem_formulation}

\vspace{-1.2mm}

\section{Approach}
This section details our radar height estimation methodology, incorporating a robust regression loss function to tackle the sparse target regression problem. Following this, the model architecture and the designed multi-task training strategy are presented.

\subsection{Radar Height Estimation}
\label{subsec:height_loss}

\begin{figure}[htbp]
\centering
     \includegraphics[width = 0.90\linewidth]{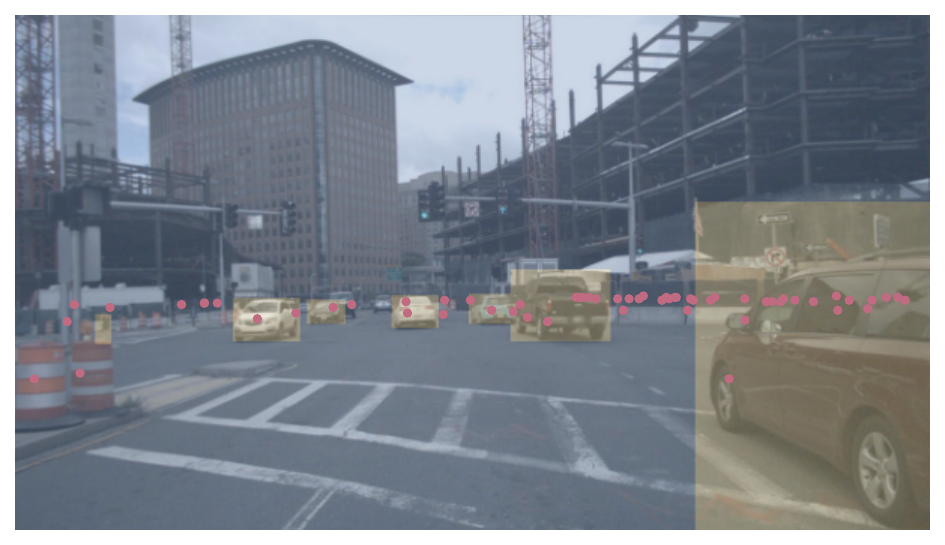}
     \caption{Ground truth height map segmentation.}
     \label{Fig:problem_exp}
     \vspace{-2.5mm}
 \end{figure}

We approach height estimation by predicting a 2D height map for every input frame. Thus, radar points in the 3D space of the current frame are first projected pixel-wise onto the image plane. These projected points are shown as red scatters in Fig. \ref{Fig:problem_exp}. We assign the correct height values to these points and store them in the ground truth height map. Radar detections that line up with 3D space bounding boxes have the same height values as the corresponding objects. Otherwise, the resulting height will remain zero. This approach leads to an excessively sparse target map without yielding effective training.
To address this, we generate a more comprehensive ground truth height map, as detailed below.

As depicted in Fig. \ref{Fig:problem_exp}, the ground truth map $H$ matches the image dimensions and is segmented into three parts. The background areas, denoted as $H_{BG}$, have a value of zero, indicating the absence of any objects. The foreground sections $H_{FG}$, where the 2D bounding boxes are found, are colored in yellow. The precise height of the object is stored in all pixels within the 2D bounding box.
Red single-pixel coordinates, denoted as $H_{RAD}$, represent radar points.
We pay particular attention to the height values associated with each radar point because these positions hold paramount significance and should not be treated equally with the other parts. There are instances where, after projection, certain points may fall within the confines of a 2D bounding box. Nevertheless, these points do not align with the actual representation of that specific object in the 3D space. This disparity has the potential to introduce confusion and misguide the model's learning process since the ground truth height values of these points should be zero rather than the height of the corresponding objects.

In the illustration of Fig. \ref{Fig:problem_exp}, the majority of the pixels in the ground truth height map carry a value of zero even after considering the height value for the 2D bounding boxes. 
These zeros become predominant in training by directly minimizing the loss between the ground truth map and the prediction. Our experiments demonstrate that this leads the model to produce a prediction map entirely filled with zeros. Thus, we propose a robust regression loss function that considers the importance of each part and pixel separately.

This work introduces a dynamic weighting factor to distinguish the importance of different pixels. This factor influences the pixel-wise loss function, enabling the network to prioritize larger discrepancies and eliminate all-zero predictions. Precisely, for a pixel situated at location $(i, j)$, the absolute difference is defined as $\Delta h = |H(i,j) - \hat{H}(i, j)|$, where $\hat{H}$ represents the predicted height map. The weighting factor is determined as $\log(\Delta h+1)$. This approach is informed by two key insights: first, it significantly penalizes inaccurate predictions, and second, the logarithmic operation improves the numerical stability. The effectiveness of the proposed weighting factor is validated into two traditional loss functions, specifically L1 and L2 loss. The weighted L1 and L2 loss are expressed as:
\begin{equation}
l_{1}(i,j)=\Delta h \times \log(\Delta h+1)
\end{equation}
\begin{equation}
l_{2}(i,j)=\Delta h^{2} \times \log(\Delta h+1)
\end{equation}

In addition to the aforementioned losses, we introduce the designed weighting factor into the \acrfull{ehl}. The loss at location $(i,j)$ is computed as follows:
\begin{equation}
\resizebox{0.91\hsize}{!}{
    $l_{EHL}(i,j) = \left\{ \begin{array}{rcl}
            \frac{1}{2} \sigma^{2} \Delta h^{2} \times \log(\Delta h+1) & \mbox{if} & \Delta h < \frac{1}{\sigma^{2}} \\
            (\Delta h- \frac{1}{2\sigma^{2}}) \times \log(\Delta h+1) & \mbox{if} & \Delta h \geq \frac{1}{\sigma^{2}}
        \end{array} \right. $
        }
\end{equation}
\noindent where the parameter $\sigma$ is a threshold, determining the transition between L1 and L2 loss. The further comparisons between these losses are detailed in Sec. \ref{subsubsec:height_result}.
\begin{figure*}
\centering
\includegraphics[width=0.9\textwidth]{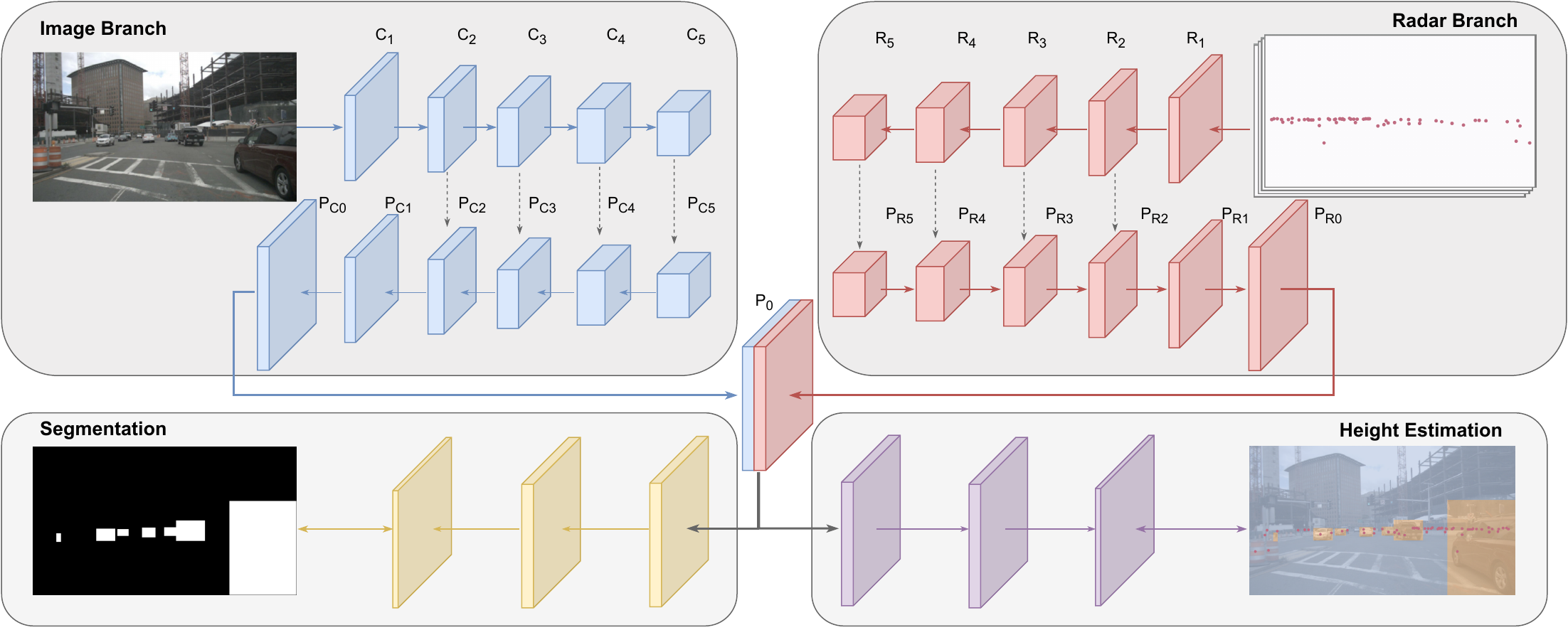}

   \caption{Model Architecture: Visual and radar images undergo individual encoding and decoding processes. Their reconstructed feature maps are concatenated and sent into the height estimation and segmentation branches for final predictions.}
\label{fig:model}

\vspace{-4mm}
\end{figure*}
For each distinct section, we calculate the average loss value, yielding $L_{BG}$, $L_{FG}$ and $L_{RAD}$. The final robust regression loss function for the height estimation is derived from a weighted combination of the three losses. Parameters $\alpha$, $\beta$, and $\gamma$ are defined to balance the importance between different parts.
\begin{equation}
    L_{reg} = \alpha L_{BG} + \beta L_{FG} + \gamma L_{RAD}
\end{equation}

\subsection{Model Architecture and Training Strategy}
To tackle the height estimation challenge, both camera and radar data are used. The radar image is generated by directly projecting radar points onto the image plane. The information carried by each radar point is stored in different channels. The camera and radar images are encoded using separate VGG16 \cite{vgg} backbones and then decoded by Feature Pyramid Networks (FPN) \cite{fpn}.
The resulting decoded feature maps, $P_{C0}$ and $P_{R0}$, encompass high-level semantic information and have the same size as the input image. These two feature maps are concatenated, denoted as $P_{0}$, and forwarded into the downstream tasks.

We subsequently introduce a multi-task training approach, including height estimation and free space segmentation, to prevent the model from making all-zeros predictions across the entire height map. In the free space segmentation aspect, the network identifies the regions where objects are present. Thus, a two-channel mask is generated for every input image based on the 2D bounding boxes. The first channel denotes the free space with a value of 1 and the occupied space with a value of 0. Conversely, free spaces in the second channel are represented with a value of 0.
The height estimation task is an advanced object segmentation task. According to the generated ground truth height map, the network has to pinpoint object locations while precisely regressing their height value. Thus, it is valuable to jointly train the network on these two tasks. The overall model architecture is illustrated in Fig. \ref{fig:model}.

The proposed network is designed to be trained end-to-end, leveraging two distinct loss functions equally tailored to specific tasks. The previously introduced loss function $L_{reg}$ is used for the height estimation, as detailed in Section \ref{subsec:height_loss}. In addition, the segmentation task is optimized using the binary cross-entropy loss, where $L_{seg}$ is computed by comparing the predicted masks to their corresponding ground truth masks.

\label{sec:approach}

\vspace{-1.2mm}
\section{Experiments}
In this section, we first describe the implementation details. Subsequently, we demonstrate the efficacy of our proposed learning-driven radar height estimation method. Finally, leveraging our methods during radar data preprocessing, we assess the performance of the CRF-Net, MCAF-Net, and DORN algorithms.
\subsection{Dataset and Implementation Details} 
We employ the nuScenes dataset for our experiments, a widely recognized dataset in autonomous driving research. 
Following the methodology described in \cite{crf, MCAF}, we derive 2D bounding boxes by projecting the provided 3D annotations onto the image plane. These 2D annotations serve as the foundation for generating the ground truth for both our segmentation and height estimation branches.
Our experiments focus on the front camera and radar data to assess the effectiveness of the proposed methods. 
The dataset is partitioned using a 3:1:1 split. To optimize the computational efficiency, we resize each visual image to $360\times 640$ pixels. 
The radar points are directly projected pixel-wise onto the image plane. This procedure results in a four-channel radar image, which is the same size as the visual image, that retains data on the \acrshort{rcs}, distance, and velocities in the x and y directions. Pixels without projected radar points are filled in with zero values in all radar channels.

The proposed models are implemented using the TensorFlow framework and are trained on the Nvidia\textsuperscript{\textregistered} Tesla A30 GPUs. We employ the Adam optimizer, initializing with a learning rate of $3e^{-4}$. In scenarios where the optimization plateaus, the learning rate is scaled down by a factor of 0.75.

The hyperparameters used in our \acrshort{ehl} are selected by validating our network thought the validation set to find the best combination of those. After validation, we find that $\sigma=3$, $\alpha=0.5$, $\beta=1$, and $\gamma=2$ is the best possible combination. 

\begin{figure}[htbp]
\centering
     \includegraphics[width = 0.8\linewidth]{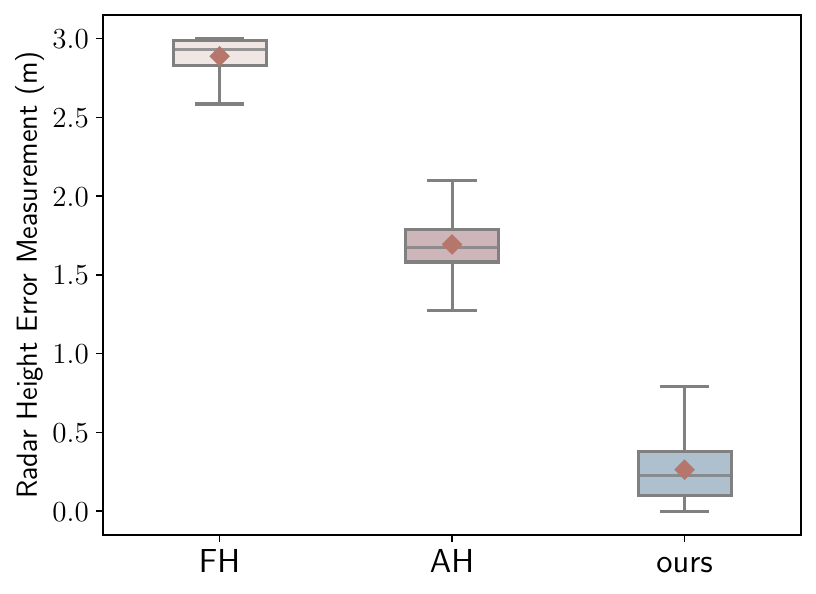}
     \caption{Radar Height Error Measurement.}
     \label{Fig: height_error}
     \vspace{-5mm}
 \end{figure}

\begin{figure*}[htbp]
     \centering
     \begin{subfigure}[b]{0.31\textwidth}
         \centering
         \includegraphics[width=\textwidth]{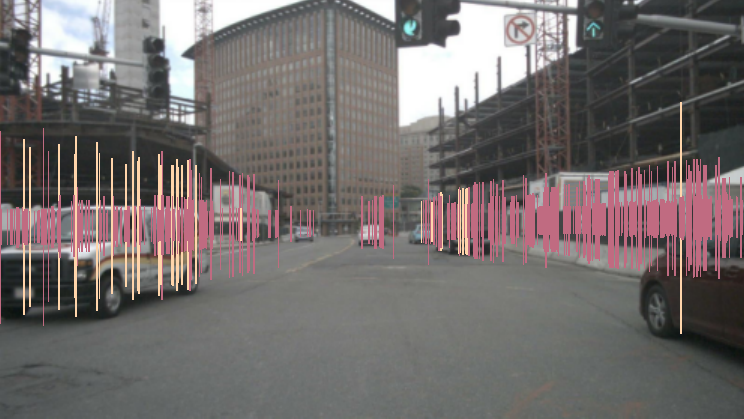}
         \caption{Fixed Height Extension}
         \label{fig:fh}
         
     \end{subfigure}
     \hfill
     \begin{subfigure}[b]{0.31\textwidth}
         \centering
         \includegraphics[width=\textwidth]{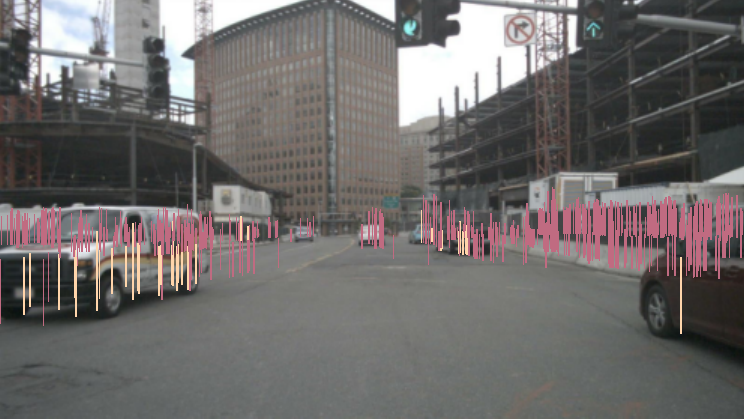}
         \caption{Adaptive Height Extension}
         \label{fig:ah}
         
     \end{subfigure}
     \hfill
     \begin{subfigure}[b]{0.31\textwidth}
         \centering
         \includegraphics[width=\textwidth]{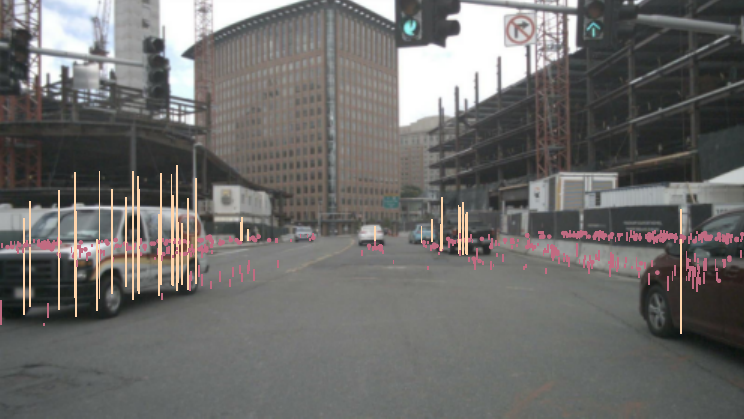}
         \caption{Ours}
         \label{fig:learnH}  
         
     \end{subfigure}
        \centering
        \caption{Qualitative comparison of height extension methods. Extended vertical lines of radar points associated with objects are colored in yellow. Others are colored in red.}
        \label{fig:qualitative}
\vspace{-5mm}
\end{figure*}

\subsection{Height Error Measurement Result}
\label{subsubsec:height_result}
In our evaluation, we delve into three analyses:

\paragraph{Compared with FH and AH} 
We evaluate our algorithm by benchmarking our learning-based height estimation technique, which employs the \acrshort{ehl}, against the traditional methods: \acrshort{fh} and \acrshort{ah}. For the \acrshort{fh} and \acrshort{ah} extensions, the height error over the test set is consistently evaluated with the methodology described in \cite{MCAF}. For our learning-based method, the height error per frame is derived by averaging the absolute differences between the predicted and ground truth maps solely at positions occupied by radar points. The comparative results are visually presented in Fig. \ref{Fig: height_error}. The learning-based method reduces the average radar height error across the test dataset from 1.69 to 0.25 meters.

The qualitative comparison of various height extension methods is depicted in Figure \ref{fig:qualitative}. Radar points associated with objects are highlighted in yellow, while those unrelated to any objects are marked in red, mostly regarded as noise. In contrast to both \acrshort{fh} and \acrshort{ah} extensions, our learning-based method exhibits superior discrimination between these two categories of radar points.
Figure \ref{fig:learnH} further illustrates the two primary advantages of our approach. Firstly, it effectively brings the predicted values of radar points with a ground truth height of 0 closer to zero, thereby mitigating the impact of noise. Secondly, our model demonstrates enhanced accuracy in predicting the heights of radar points associated with objects. These advantages culminate in the production of higher-quality radar data.

\paragraph{Evaluating the effectiveness of EHL}
We evaluate the performance of various models trained using different loss functions: L1, L2, Weighted L1 (WL1), Weighted L2 (WL2), Huber Loss (HL), and \acrshort{ehl}. Besides assessing the Radar Height Error (RHE), we also consider the Box Height Error (BHE), which is determined by averaging the absolute differences between the prediction and the ground truth map. Additionally, we differentiate the radar points based on their association with objects. Specifically, we calculate the average radar height error for points linked with an object, where height is not 0, as RHE$_{\neq 0}$ and for points unassociated with any object, where height is 0, as RHE$_{= 0}$. The outcomes are presented in Table \ref{table:error_for_loss}. The model trained with proposed \acrshort{ehl} demonstrates fewer errors than those trained with other loss functions. The significance of the weighting factor becomes evident when comparing results from L1 and WL1 loss.
Specifically, the model trained using the L1 loss consistently predicts all zeros. However, WL2 underperforms compared to L2, as including the weighting factor amplifies the impact of outliers.
\begin{table}[ht]
\centering
\begin{tabular}{l||cccc}
\hline
Loss & BHE(m) & RHE(m) & RHE$_{\neq 0}$(m) & RHE$_{= 0}$(m) \\ \hline
L1   & N/A   &   N/A       & N/A &  N/A \\
L2           & 0.34   & 0.41     & 0.71   &    0.41    \\
HL          & 0.25   & 0.26    & 0.68   &    0.24    \\
WL1       &  0.29  &   0.29      &  0.68   &   0.28    \\
WL2        &  0.70   &  0.61     &  0.73     & 0.61 \\
EHL      &  \textbf{0.23}    &  \textbf{0.25}         &  \textbf{0.66}  & \textbf{0.24}        \\ \hline
\end{tabular}
\caption{Height error comparison between different losses.}
\label{table:error_for_loss}
\vspace{-3mm}
\end{table}
\paragraph{Importance of the segmentation branch} We further evaluated the models trained solely minimizing \acrshort{ehl}, focusing only on the height estimation branch. This resulted in a RHE increase to 0.51 meters, which falls behind compared to models trained jointly on both tasks. Additionally, it's noteworthy that single-task models, when trained with other loss functions, consistently yielded all-zero predictions.

\subsection{Perception Tasks Evaluation} After successfully training the height estimation model, we retain the height values of radar points across all frames for subsequent processing. We propose two methodologies for generating the final projected radar channels. The \emph{Direct Method} involves substituting the height values of each radar point with those predicted by our model. In contrast, the \emph{Filter Method} entails the removal of radar points whose predicted heights are below 0.5 meters in each frame. This filtering is based on the premise that the objects of interest typically exceed 0.5 meters in height. The radar image is subsequently constructed using the remaining points.

The experimental setups in these studies adhere to the protocols established in the original papers \cite{crf, MCAF, Dorn_radar}.
\subsubsection{Object Detection Evaluation}
In this phase, we retrain both the CRF-Net and MCAF-Net using our enhanced radar data. Model performance is assessed using the widely accepted \acrshort{map} metric. The outcomes of this evaluation are summarized in Table \ref{table:object_detect}.

\begin{table}[ht]
\centering
\begin{tabular}{l||cccccc}
\hline
  & mAP & Night mAP & Rain mAP \\ \hline
MCAF-Net \cite{MCAF} & 47.70\%    &  49.77\%         & 44.91\%    \\
Ours (Direct)       & 47.98\%    & \textbf{50.33 \%}         & 44.99\%        \\
Ours (Filter)      & \textbf{48.27\%}    & 49.89\%          & \textbf{ 45.02\%   }      \\
\hline
CRF-Net \cite{crf} & 43.83\%    &  46.12\%         & 41.04\%    \\
Ours(Direct)      &44.29\%   & 46.61\%   &41.69\%   \\
Ours(Filter)  &  \textbf{44.88\%}  &  \textbf{46.65\%}  &  \textbf{41.90\%}\\
\hline
\end{tabular}
\caption{Comparison of the performance using different radar preprocessing methods for CRF-Net and MCAF-Net.}
\vspace{-3mm}
\label{table:object_detect}
\end{table}

\subsubsection{Depth Estimation Evaluation}
For depth estimation, we replace the extended height values of each radar point with the heights regressed by our model. We denote the set of 2D pixels with ground truth Lidar depth values as $\Omega$. The predicted and ground truth depth maps are represented by $\hat{d}$ and $d_{gt}$, respectively. The evaluation of model performance involves computing the Mean Absolute Error (MAE), Root Mean Square Error (RMSE), Absolute Relative Error (AbsRel), and the $\delta _{n}$ threshold. The metrics are defined in Table \ref{table:metrics}, and the corresponding results are presented in Table \ref{table:dorn}.

\begin{table}[ht]
\centering
\begin{tabular}{l||cccc}
\hline
 & Definition\\ \hline
MAE  &  $\frac{1}{|\Omega|}\sum_{x\in \Omega} |\hat{d}(x)-d_{gt}(x)|$      \\ 
RMSE    &   $ (\frac{1}{|\Omega|}\sum_{x\in \Omega} |\hat{d}(x)-d_{gt}(x)|^{2})^{1/2} $  \\
AbsRel       & $\frac{1}{|\Omega|}\sum_{x\in \Omega} |\hat{d}(x)-d_{gt}(x)|/d_{gt}(x)$ \\
$\delta_{n}$ threshold   &   $\delta_{n}=|\{\hat{d}(x): max(\frac{\hat{d}(x)}{d_{gt}(x)}, \frac{d_{gt}(x)}{\hat{d}(x)})< 1.25^{n}\}|/|\Omega|$        \\
 \hline
\end{tabular}
\caption{Metrics definition for depth estimation task.}
\label{table:metrics}
\vspace{-3mm}
\end{table}

\begin{table}[ht]
\centering
\begin{tabular}{l||cccc}
\hline
 & MAE $\downarrow$ & RMSE $\downarrow$ & AbsRel $\downarrow$ & $\delta_{1}$ $\uparrow$\\ \hline
DORN \cite{Dorn_radar}  & 2.432   &   5.304      & 0.107 &  0.890 \\
Ours (Direct)          & \textbf{2.381}  & 5.287   & \textbf{0.101}  &    0.892   \\
Ours (Filter)        & 2.395  & \textbf{5.280}   & 0.102  &   \textbf{0.897}   \\
 \hline
\end{tabular}
\caption{Comparison of the performance using different radar preprocessing methods for the DORN algorithm.}
\label{table:dorn}
\vspace{-4mm}
\end{table}

From the tabular results, it is evident that the utilization of our high-quality radar data enhances algorithmic performance, underscoring the significance of our work.
\label{sec:experiments}

\vspace{-1.2mm}

\section{Conclusion}
This study introduces a multi-task learning framework for the height estimation of radar points, featuring a novel robust regression loss function with a weighting factor tailored for sparse target regression. Our model is further enhanced by integrating a free space segmentation task, enabling effective differentiation between foreground and background elements. Notably, our approach significantly lowers the RHE from 1.69 to 0.25 meters, surpassing the performance of the AH extension. The height estimation task proposed here also serves as a vital preprocessing step to refine radar data for subsequent perception tasks. 
It adeptly denoises radar data by filtering out points with predicted heights below a certain threshold, thereby yielding high-quality radar data. 
This refined data substantially enhances the efficacy of 2D object detection and depth estimation algorithms, where the radar points are projected onto the image plane. 
Looking ahead, we plan to utilize our high-quality radar data in various perception tasks within sensor fusion frameworks, such as 3D object detection.
Within this context, each individual point can be expanded into a pillar, using the height value extracted from our framework rather than the conventional method of employing a fixed-size pillar, to demonstrate its versatility and utility in advanced applications.
\label{sec:conclusion}
\vspace{-1.5mm}


\section{Acknowledgement}
Research leading to these results 
has received funding from the EU ECSEL Joint Undertaking under grant agreement n° 101007326 (project AI4CSM) and from the partner national funding authorities the German Ministry of Education and Research (BMBF).
\label{sec:acknowledgement}





\vspace{-1.5mm}
\bibliographystyle{IEEEtran}
\bibliography{references}

\end{document}